\documentclass{article}

 \usepackage[final]{nips_2018}

\usepackage[utf8]{inputenc} 
\usepackage[T1]{fontenc}    
\usepackage{hyperref}       
\usepackage{url}            
\usepackage{booktabs}       
\usepackage{amsfonts}       
\usepackage{nicefrac}       
\usepackage{microtype}      
\usepackage{graphicx}
\usepackage{subfig}
\usepackage{booktabs}
\title{Population-aware Hierarchical Bayesian Domain Adaptation}

\author{
Vishwali Mhasawade$^{1}$, Nabeel Abdur Rehman$^{1}$, Rumi Chunara$^{1,2}$
\\
$^{1}$\normalfont Department of Computer Science and Engineering, Tandon School of Engineering\\
$^{2}$\normalfont Department of Biostatistics, College of Global Public Health\\
New York University \\
\texttt{\{vishwalim, nabeel, rumi.chunara\} @nyu.edu}
  }
  
\begin{document}
\maketitle
\begin{abstract}

Population attributes are essential in health for understanding who the data represents and precision medicine efforts. Even within disease infection labels, patients can exhibit significant variability; ``fever'' may mean something different when reported in a doctor's office versus from an online app, precluding directly learning across different data sets for the same prediction task. This problem falls into the domain adaptation paradigm. However, research in this area has to-date not considered who generates the data; symptoms reported by a woman versus a man, for example, could also have different implications. We propose a novel population-aware domain adaptation approach by formulating the domain adaptation task as a multi-source hierarchical Bayesian framework.
The model improves prediction in the case of largely unlabelled target data by harnessing both domain and population invariant information.

\end{abstract}

\section{Introduction}
Standardization in clinical case definitions is a significant challenge. This is becoming more pertinent as the number and types of places, modes of data collection and populations generating data are expanding (from clinical data to healthworker-facilitated data wherein healthworkers visit individuals' houses, record symptoms and take specimens, to citizen-science studies in which participants report symptoms from home and mail in or submit specimens \cite{goff2015surveillance,fragaszy2016cohort}) making infection prediction based on a specific syndromic case definition (set of symptoms) challenging. Moreover, it’s extremely rare for data from different studies to be collected in the exact same mode, context and from the same type of population. Therefore symptoms (features) can mean different things; “fever” may mean something different reported to a doctor than at home through a smartphone app \cite{ray2017predicting,rehman2018domain}. Furthermore, how young people report may be different from how older people report symptoms. These differences in the data collection as well as the variance in the demographic distributions of the different datasets make the important problem of predicting infection based on syndromic case definitions challenging. 

Early work has shown that public health collection methods can be conceptualized as domains, and domain adaptation can be useful for prediction from symptom data sets obtained via these different modes \cite{rehman2018domain}. Beyond this, to the best of our knowledge no work has addressed the issue of domain differences in health data while also accounting for population attributes (work has also only focused on improving prediction in a target data set via the use of a single source, whereas the work here uses multiple sources from different domains). Incorporation of population structure has not been explored extensively, though in health practice and research attributes of the people contributing the data (here we consider population demographics like age, gender) are commonly available, and there are shared characteristics within these groups \cite{saria2010learning}. While increasing representation granularity by increasing the number of classes can help, ad hoc discretization into fixed sets can limit ability to model instance-specific variability. Therefore hierarchical approaches have been used (but not yet for domain adaptation); for example Dirichlet processes have been used to allow sharing of mixture components in time-series data, generating global and individual topic parameters \cite{saria2010learning}.


Hierarchical approaches have primarily been developed in natural language processing, and use Bayesian priors to tie parameters across multiple tasks \cite{evgeniou2005learning}. In such methods, each domain has its own domain-specific parameter for each feature which the model links via a hierarchical Bayesian global prior instead of a constant prior. This prior encourages features to have similar weights across domains, unless there is good contrary evidence. Hierarchical Bayesian frameworks are a more principled approach for transfer learning, compared to approaches which learn parameters of each task/distribution independently and smooth parameters of tasks with more information towards coarser-grained ones \cite{carlin2010bayes,mccallum1998improving}. An undirected Bayesian transfer hierarchy has been used to jointly model the shapes of different mammals \cite{elidan2012convex}.  While we build on this idea of hierarchical modeling for domain adaptation, here we go further to explicitly model population attributes via hierarchical structure. Also, given that health-related data sets can be collected in many different ways and from varied population samples, we explicitly consider a multi-source situation using empirical information about the included population in multiple studies to contribute to learning the model posterior and improve transfer of information to a new population and domain, with limited infection labels. 
\section{Data}
Each dataset includes symptoms from individuals, laboratory confirmation of type of respiratory infection virus they had (if any), as well as the age and gender of the person as example basic population attributes. We group these attributes into categories (gender as male/female, age as 0-4 years, 5-15 years, 16-44 years, 45-64 years and 65+ years) \cite{rehman2018domain}. \textbf{GoViral} data was collected from volunteers who self-reported symptoms online and also mailed in bio-specimens for laboratory confirmation of illness in New York City. It consists of 520 observations out of which 291 had positive laboratory results \cite{goff2015surveillance}. \textbf{FluWatch} consists of 915 observations (567 positive cases of flu) of volunteers in the United Kingdom. These two datasets belong to the ``citizen science'' domain \cite{fragaszy2016cohort, rehman2018domain}. \textbf{Hong Kong} consists of 4954 observations (1471 positive cases of flu) collected by healthworkers in Hong Kong \cite{cowling2010comparative}. The \textbf{Hutterite} data is composed of 1281 observations (787 positive cases of flu) of colonies in Alberta, Canada sampled by nurses \cite{loeb2010effect}. The high variability in attribute distributions in these real-world datasets is illustrated in Figure \ref{fig:model+demo}(i).

\section { Methods}

\begin{figure}
\centering
\begin{minipage}{.56\textwidth}
  \centering
  \includegraphics[trim=5cm 2cm -2cm 1.5cm,scale = 0.3]{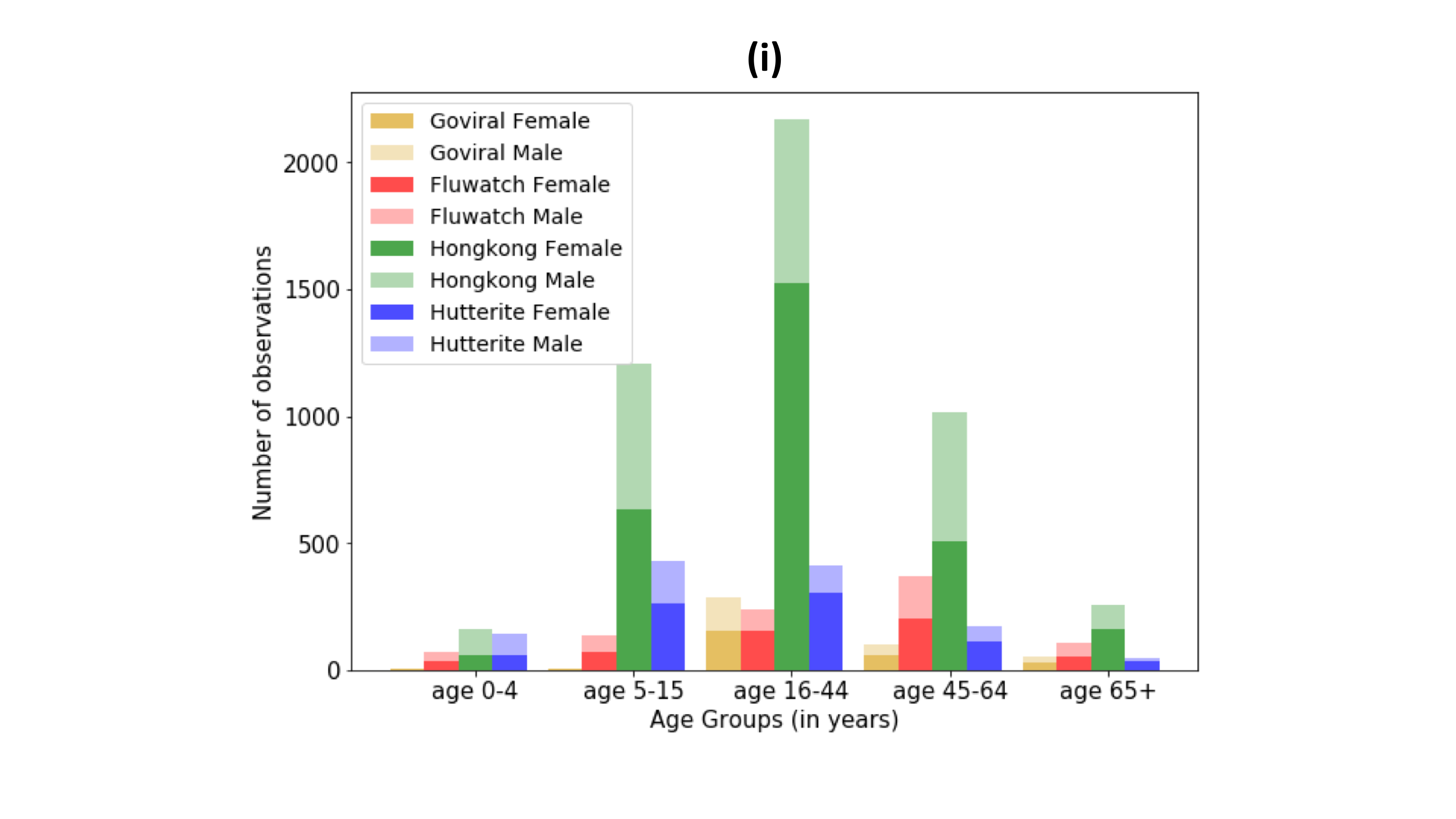}
  \captionsetup{justification=centering}
  \label{fig:demo}
\end{minipage}%
\begin{minipage}{.6\textwidth}
  \centering
  \includegraphics[trim= 3.5cm 0cm 6cm 1.5cm, scale=0.29]{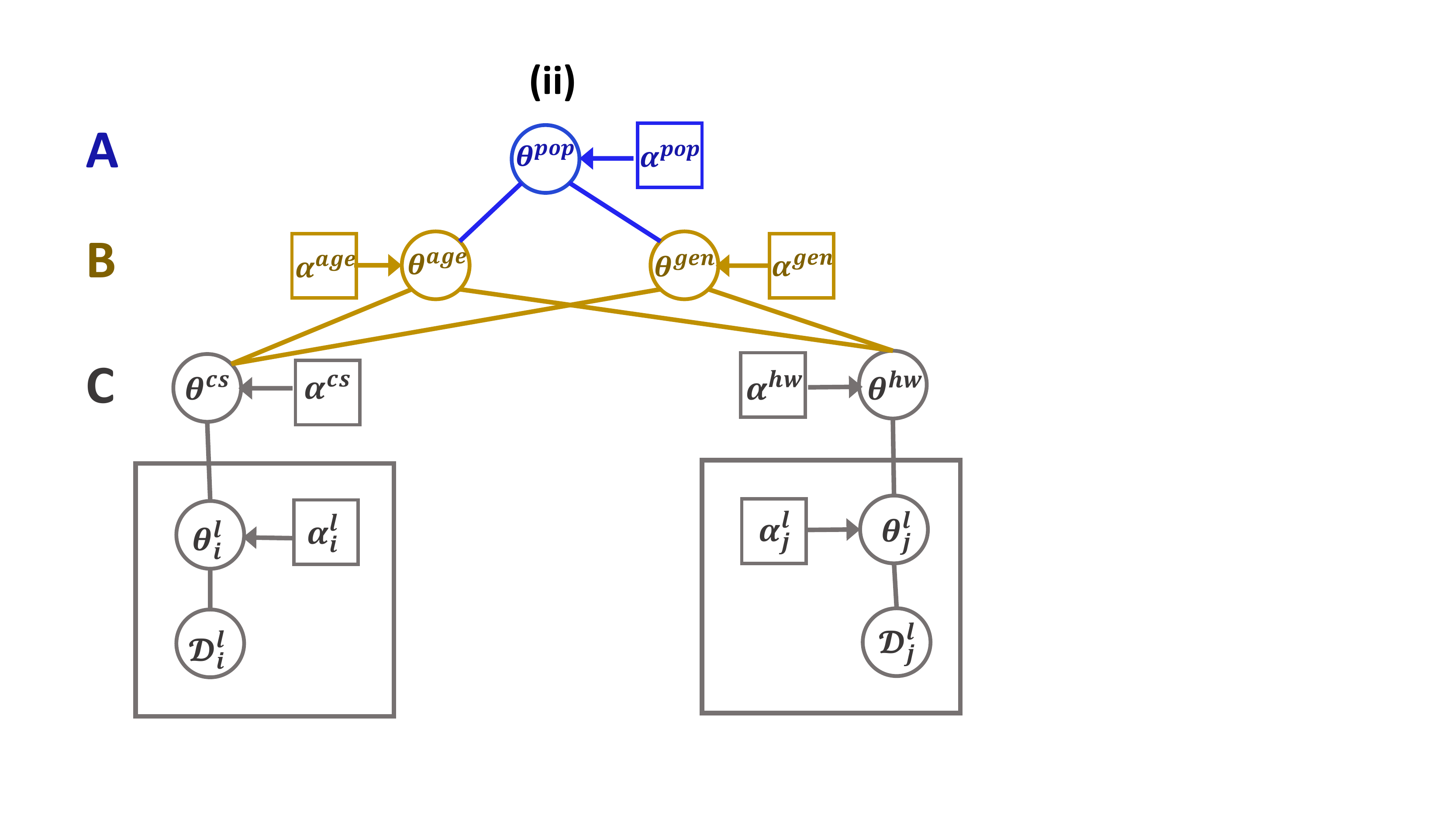}
  \captionsetup{justification=centering}
  \label{fig:model}
  
\end{minipage}
\vspace{-0.425cm}
\caption{(i) Demographic distributions of the datasets. (ii) The population-aware hierarchical model; $\theta$ parameters at different nodes, $\mathcal{D}$ different data sets and $\alpha$ the priors. ii(A): Root level that represents invariant information across all data, ii(B): population parameters and information invariant to population-attributes (here, age and gender), ii(C): data set and domain-specific parameters and information (here, for $i$ citizen science (CS) and $j$ healthworker facilitated (HW) domain datasets). }
\label{fig:model+demo}
\end{figure}




\textbf{Overall Undirected Hierarchical Multi-source Bayesian Approach} In this framework, the lowest level of the hierarchy represents the datasets (within each domain (in our case, collection mode), \(l\in \mathcal{L}\), for each of which we have data $\mathcal{D}^l$ as shown in Figure \ref{fig:model+demo}. As in all Bayesian problems, the dataset parameters $\theta^{l}$ should represent the data $D^{l}$ well. Here, $\theta^{l}$ are influenced by the domain-specific parameters ($\theta^{s}$); $\theta^l$ are generated according to $P(\theta^{l}|\theta^{s})$, where \(s \in \mathcal{S}\) is the domain. In the undirected population-aware hierarchical model we allow the domain specific parameters to have multiple parents and learn all parameters simultaneously. Accordingly, the domain parameters are generated according to the distribution $P(\theta^{s}|\theta^{g}, \theta^{a})$. Here, we explicitly include $\theta^{a}$ to represent the population parameters (here $a \in \mathcal{A}$ for the different age group categories, and similarly for genders $\theta^{g}$ where $g \in \mathcal{G}$. The population parameters $\theta^{g}$ and $\theta^{a}$ have the root parameter $\theta^{pop}$ as the parent, which represents invariant information across all of the datasets, classes and population attributes, $P(\theta^{pop}|\theta^{par(pop)}) \equiv P(\theta^{pop})$. 
Then, the joint distribution accounting for all of these data and parameters is:
$P(\mathcal{D}, \theta) =   \prod_{l \in \mathcal{L}} P(\mathcal{D}^{l}|\theta^{l}) \times 
    \prod_{l \in \mathcal{L}} P(\theta^{l}|\theta^{s})  \times
    \prod_{s \in \mathcal{S}} P(\theta^{s}|\theta^{a},\theta^{g}) 
    \times \prod_{a \in \mathcal{A}}
    P(\theta^{a}|\theta^{pop}) \times
    \prod_{g \in \mathcal{G}}
    P(\theta^{g}|\theta^{pop}) \times P(\theta^{pop})$.


    
\textbf{Hierarchy Priors} For all parameters we use independent priors that are computed based on symptom predictivity for each age group and gender. 
The inclusion of data dependent priors in Bayesian learning has been explored to incorporate domain knowledge into the posterior distribution of parameters \cite{darnieder2011bayesian}. 
For population-aware modeling, data-informed prior distributions are important because the distributions from each dataset are particular to the study, and thus capturing this information adds more information to the analysis than improper or vague priors (e.g. for a sample wherin one demographic group is under-represented), also motivates the multiple parents in the hierarchy.
In contrast, using just the root prior for estimating the posterior ignores the demographic information available.  
Therefore, we use an empirical Bayes approach to specify weakly informative priors, centered around the estimates of the model parameters \cite{van2017prior}. Root parameters are centered on the cumulative data since the root parameter captures domain invariant information. 

\textbf{Model Steps} First, we use a probabilistic framework to jointly learn each parameter based on all levels of the hierarchy. We use a maximum a-posteriori parameter estimate instead of the full posterior for the joint distribution, which would be computationally intractable. We use a formulation, proposed in \cite{elidan2012convex} that is amenable to standard optimization techniques, resulting in the objective:
\vspace{-0.1cm}
\begin{equation}
   F_{objective} = -\sum_{d \in D} \bigg[ \sum_{j} (f_j  + \lambda) + \theta_{j}^{d}
   - \log \sum_{k} \exp(\theta_{k}^{d})\bigg] 
   + \beta\sum_{l \in Nodes} \mbox{Div}(\theta^{l}, \theta^{par(c)})
    \label{eqn:obj}
\end{equation}

\vspace{-0.1cm}

\begin{table*}[b]
\centering
\caption{AUC comparison for flu prediction task (with 20\% labelled data from the target dataset).}
\begin{tabular}{l c c c c }
\toprule
	 & Goviral & Fluwatch & Hongkong & Hutterite \\
\midrule
    Target Only & 0.652 & 0.590 & 0.890 & 0.749 \\
    Logistic Regression & 0.681 & 0.461 & 0.882 & 0.748  \\
    FEDA (Only symptoms) & 0.675 & 0.521 & 0.900 & 0.726 \\
    FEDA+p (With demographics) & 0.693 &  0.612 & 0.914 & 0.824 \\
    Hierarchical (Only symptoms) & 0.685 & 0.486 & 0.889 & 0.719 \\
    Hierarchical+p (With demographics) & \textbf{0.710} & \textbf{0.627} & \textbf{0.918} & \textbf{0.827} \\
    
\bottomrule
\end{tabular}
\label{tab:results}
\end{table*}

For datasets $d$, $\theta_{j}^{d}$ denotes the parameter for  symptom $j$. From a specific dataset's parameter space, $k$ denotes each symptom. $f_{j}$ is a statistical measure of the symptom $j$ in the dataset. In this case the statistical measure is the proportion of the particular symptom resulting in a positive cold/flu test (i.e. the positive predictive value). $Nodes$ is the set of all nodes in the hierarchy (here, $\mathcal{L} \cup \mathcal{S} \cup \mathcal{A} \cup \mathcal{G}$). We consider the case of the parameters belonging to a multinomial distributions and consider the log representation. Regularizing parameter $\lambda$ was chosen to be 1 to allow Laplacian smoothing \cite{mccallum1998improving}. The function $\mbox{Div}(\theta^{l},\theta^{par(c)})$ is a divergence (L2 norm used) over the child and the parent parameters that encourages child parameters to be influenced by parent parameters, and allows a child parameter to be closely linked to more than one parent. The weight $\beta$ represents the influence balance between node parameters and node parent parameters. Based on hyperparameter tuning, a value of 0.2 for $\beta$ was used in all experiments. For optimization of the objective function, we use Powell's method \cite{fletcher1963rapidly}.

Second, we learn the influence of the each hierarchical levels for a particular dataset; this is done to enable the model to give more weight to one level of the hierarchy when needed. In other words, how much demographic-invariant or domain-invariant information is needed depends upon how much information is in a given dataset. The reason for learning the weights for the different levels for each dataset independently is that each dataset would require different amounts of information from the demographic-specific and the domain-specific parameters, depending upon the demographic distribution of the sample in that dataset as well as the collection mode. Different weights for the domains stems from the fact that each symptom could mean something different across collection mode; `fever' when individually reported through citizen efforts has a different predictive power as when it is collected in a standardized way collected via a healthworker. We use a simple logistic regression for this optimization.



\section{Experiments}

 \begin{table*}[t]
 \centering
 \caption {AUC scores across increasing proportions of training data (best two models).}
 \begin{tabular}{l @{\hspace{1.5\tabcolsep}} c @{\hspace{1.5\tabcolsep}} c@{\hspace{1.5\tabcolsep}} c @{\hspace{1.5\tabcolsep}} c @{\hspace{1.5\tabcolsep}} c @{\hspace{1.5\tabcolsep}} c @{\hspace{1.5\tabcolsep}} c @{\hspace{1.5\tabcolsep}} c}
 \toprule
     {Proportion} &
       \multicolumn{2}{c}{10\%} &
       \multicolumn{2}{c}{15\%} &
       \multicolumn{2}{c}{20\%} &
       \multicolumn{2}{c}{25\%}\\
    \midrule
     Dataset & FEDA+p & Hier+p & FEDA+p & Hier+p &FEDA+p & Hier+p & FEDA+p & Hier+p\\
    \midrule
     Goviral & 0.670 & 0.671 & 0.634 & 0.664 & 0.693 & 0.71 &0.664 & 0.690 \\
     Fluwatch & 0.724 & 0.576  & 0.718 & 0.699 & 0.612 & 0.627 & 0.757 & 0.710 \\
     Hongkong & 0.896 & 0.911 & 0.971 & 0.984 & 0.914 & 0.918 & 0.969 & 0.940 \\
     Hutterite &0.742 & 0.785  & 0.873 & 0.880 & 0.824 & 0.827 & 0.879 & 0.800 \\
     \bottomrule
   \end{tabular}
   \label{tab:comparision}
 \end{table*}


As motivated, we consider the case of transferring information from multiple source data sets from different domains to a largely unlabelled target dataset. In each experiment three source datasets are used in entirety along with a small amount of labelled target data. Area under the ROC curve (AUC) metric is used to assess performance. While each of the datasets have a varied composition in terms of total number of observations and population demographics, we choose to use them all without any pre-processing, as these demonstrate real data set differences and will indicate model performance in such real-world situations. We compare results to five methods to specifically examine the benefit of the (1) hierarchical structure and (2) incorporation of population attributes: Target only (\textbf{Target}), Logistic Regression (\textbf{LR}), Frustratingly Easy Domain Adaptation, which is noted for extreme simplicity and was used previously on symptom data \cite{daume2009frustratingly,rehman2018domain}, just with symptoms (\textbf{FEDA}),  and with both symptoms and population attributes (\textbf{FEDA+p}), Undirected Hierarchical Bayesian Domain adaptation without population attributes (\textbf{Hier}) and with population attributes (\textbf{Hier+p}).


\section{Results and Discussion}
Of the methods compared, Target and LR have the poorest performance (Table\ref{tab:results}). This makes sense, as a target-only model doesn't incorporate any information from other domains or populations. And, LR doesn't account for any population attributes. These methods also perform worse than the domain adaptation methods (FEDA and Hier). This indicates that there is domain-specific structure to the data. Finally, the methods that do account for population attributes perform the best. Generally the Hier+p method performs the best; this was studied more based on amount of labelled training data available (below). We also examined the learned parameters, finding that they are intuitive and generally interpretable. If a particular demographic attribute has information about the symptom predictivity, then the weight for it's influence is high. For example, in the case of the Goviral data, we find that the weights of the gender parameters are close to each other, while the proportion of males to females who were positive for flu/colds is also close to 1. In situations of low amount of training data (e.g. in GoViral, for which there are no observations in the age group of 5-15), correspondingly the influence weight of those categories is low.

The amount of Target data used in Table \ref{tab:results} was chosen based on how performance varies by proportion labels available (Table \ref{tab:comparision}). Results show that Hier+p generally worked better with under 25\% labels indicating this approach is particularly suitable in cases of very limited training data. This makes sense, as when more labels are available, more information about features is available and less reliance on population-invariant information is needed. The model harnesses population invariant information from the other data sets (multi-source learning) and domain adaptation to improve prediction on a target when very little feature-specific information is available.  Given these findings, we are interested in developing a generalizable framework for understanding how domain and population distribution differences affect results (e.g. Fluwatch poorer results at 10-15\% target labels).

\section*{Acknowledgments}
This work was supported in part by National Science Foundation grants 1643576 and 1551036.








\bibliographystyle{plain}
\bibliography{nips_ref}

\end{document}